\documentclass[letterpaper, 10 pt, conference]{ieeeconf}  

\IEEEoverridecommandlockouts                              

\usepackage{graphicx}
\usepackage{subfloat}
\usepackage{csquotes}
\usepackage{svg}
\usepackage{tabularx, multirow, multicol}
\usepackage{amsmath}
\usepackage{amssymb}
\usepackage{booktabs}
\usepackage{subcaption}
\usepackage{lettrine}
\usepackage{diagbox}

\usepackage[breaklinks,hidelinks]{hyperref}

\usepackage[capitalize]{cleveref}
\crefname{figure}{Fig.}{Figs.}
\crefname{section}{Sec.}{Secs.}
\crefname{table}{Tab.}{Tabs.}
\crefname{equation}{Eq.}{Eq.}

\newcommand{\etal}{\emph{et al}}

\title{\LARGE \bf
Region of Interest Loss for Anonymizing Learned Image Compression
}

\author{
    Christoph Liebender$^{1,2}$ \and Ranulfo Bezerra$^{1}$ \and Kazunori Ohno$^{1}$ \and Satoshi Tadokoro$^{1}$
    \thanks{$^{1}$Graduate School of Information Sciences, Tohoku University, Japan.}
    \thanks{$^{2}$Christoph Liebender is with the robotics group of Julius Maximilian University Würzburg, Germany. This paper originated during his research stay at Tohoku University. {\tt\scriptsize christoph.liebender@uni-wuerzburg.de}}
    \thanks{This work was partially supported by the robotics group of the Julius Maximilian University Würzburg, the Innovation and Technology Commission of the HKSAR Government under the InnoHK initiative, and this research was performed by the commissioned research fund provided by F-REI (JPFR23010101).}
}

\begin{document}

\maketitle
\thispagestyle{empty}
\pagestyle{empty}

\begin{abstract}
    The use of AI in public spaces continually raises concerns about privacy and the protection of sensitive data. 
    An example is the deployment of detection and recognition methods on humans, where images are provided by surveillance cameras.
    This results in the acquisition of great amounts of sensitive data, since the capture and transmission of images taken by such cameras happens unaltered, for them to be received by a server on the network.
    However, many applications do not explicitly require the identity of a given person in a scene;
    An anonymized representation containing information of the person's position while preserving the context of them in the scene suffices.
    We show how using a customized loss function on region of interests (ROI) can achieve sufficient anonymization such that human faces become unrecognizable while persons are kept detectable, by training an end-to-end optimized autoencoder for learned image compression that utilizes the flexibility of the learned analysis and reconstruction transforms for the task of mutating parts of the compression result.
    This approach enables compression and anonymization in one step on the capture device, instead of transmitting sensitive, nonanonymized data over the network.    
    Additionally, we evaluate how this anonymization impacts the average precision of pre-trained foundation models on detecting faces (MTCNN) and humans (YOLOv8) in comparison to non-ANN based methods, while considering compression rate and latency.
\end{abstract}

\section{Introduction}
\label{sec:intro}

\begin{figure}[t]
    \centering
    \includegraphics[width=\linewidth]{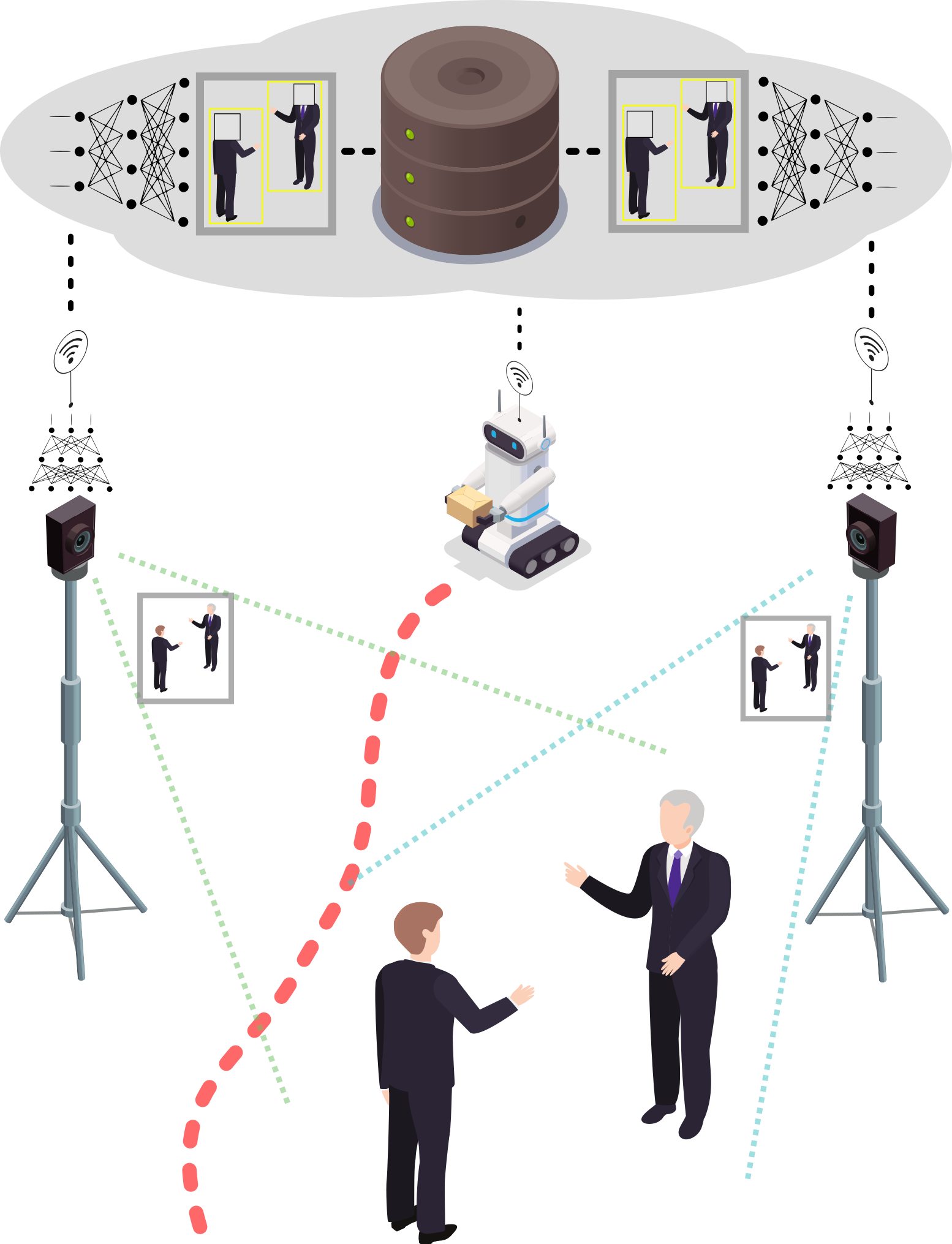}
    \caption{Depiction of example usage application: collision avoidance system with humans in the action space of mobile robots. Cameras use our method to anonymize and compress images of the scene, for a server to first decode the data and then to further process it by detecting the position of individuals (yellow bounding boxes) while their biometric information is obfuscated (boxes on faces) by the compression / decompression process. Robots rely on the detected positions plan their path in the shared space.}
    \label{fig:concept}
\end{figure}

\lettrine{\textbf{U}}{tilizing} cameras to obtain visual information of a given scene is a fundamental part of remote sensing and at the core of many applications.
Digital twin generation of a scene is one of these applications.
In this task, image data of many cameras can be used to digitally replicate a 3D scene with its spatial and semantic information.
An emerging use case for digital twin generation are industry applications, where a manufacturing site or warehouse is monitored by such cameras.
With a digital twin of such a site, mobile robots may be added to the industry process much easier, since task and route planning is aided by an accurate digital twin.
Shown in \Cref{fig:concept}, this generation of semantic information becomes even more important as soon as humans join the action space of those mobile robots, as robots then need to adapt to the actions of humans, which requires the detection of both positions and actions of them in the scene.
\par To generate an accurate digital twin, multiple viewpoints of the scene are necessary to reduce occlusions.
This yields a requirement for many cameras, that then transmit their image data to a processing server.
As it turns out, many cameras sharing a network to send their high-resolution image data quickly reach the bandwidth limit of that network, yielding a requirement of some form of image or video compression.
This pre-processing before transmission requires computational resources present on the camera, turning it into an edge-computing device.
\par Since these edge-devices now share a network, we must consider the security implications that come with this approach.
The transmitted image data contains potentially sensitive information, especially in public environments.
Even a network that is considered trusted bears a chance that an attacker may recover the data that is transmitted over it.
Implementations of such an approach must ensure that sensitive data is kept private.
\par When image data is collected in public spaces, privacy is not the only consideration to make, as the continuous recording and tracking of individuals records biometric information that is not necessary for a task like location and action detection.
This yields an additional requirement of sufficient anonymization of the data.
In fact, this matter has reached as high a level as the European Parliament:
the so-called \emph{Artificial Intelligence Act} as proposed in 2021 puts \textquote{real-time remote biometric identification systems in publicly accessible spaces} in the category of \textquote{prohibited AI practices}, the highest prohibition category \cite{EU2023}.
\par However, edge-side anonymization requires edge devices to have the computational resources to both employ anonymization and compression as \emph{seperate} tasks.
Current ANN-based anonymization techniques have not been developed with the aspect of platform constraints in mind; That is: edge devices like those that we use for this research cannot run state of the art anonymization models since their weights do not fit into device memory.
As an example, similar approaches like that of Yang \etal. \cite{Yang2023} and More \etal. \cite{More2018} use graphics cards with 11GB of memory for their experiments, while we manage to reduce this requirement down to 4GB of a Jetson Nano.
Furthermore, edge devices may not come with hardware implementations of the employed codecs, adding to the latency of ANN anonymization.
Ideally, the computational load of edge-side anonymization is shifted towards the server-side, while there is no need in trusting a vulnerable communication channel.
\begin{figure}[t]
  \centering
   \subfloat[Original, $512 \times 512$]{\includegraphics[width=.465\linewidth]{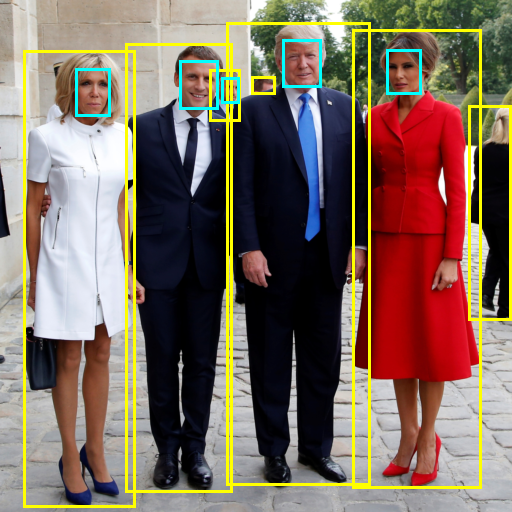}\label{fig:image-comparison-original}} \hfill
   \subfloat[Ours @ $0.1973$ bpp]{\includegraphics[width=.465\linewidth]{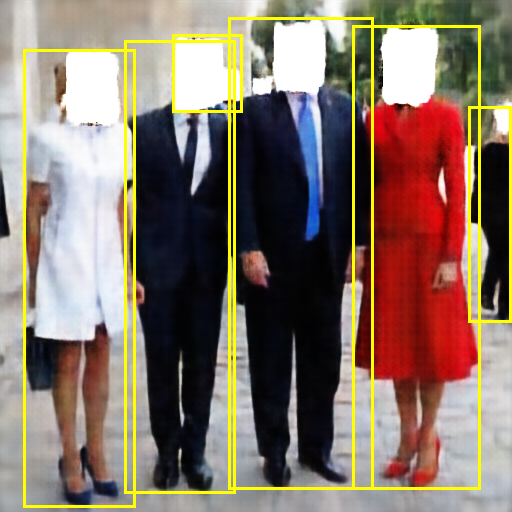}\label{fig:image-comparison-ours}}\\
   \subfloat[AV1 @ $0.2018$ bpp]{\includegraphics[width=.465\linewidth]{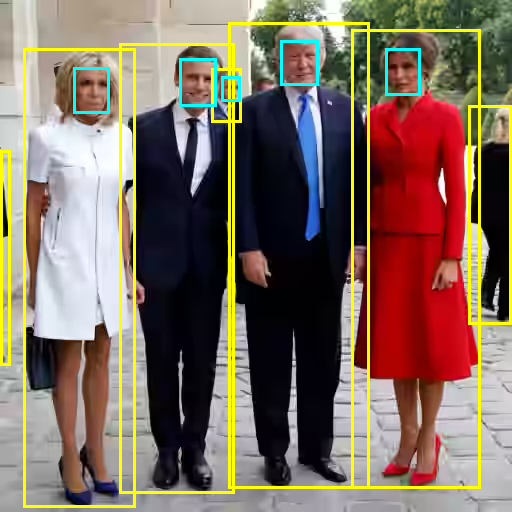}\label{fig:image-comparison-av1}} \hfill
   \subfloat[JPEG @ $0.2059$ bpp]{\includegraphics[width=.465\linewidth]{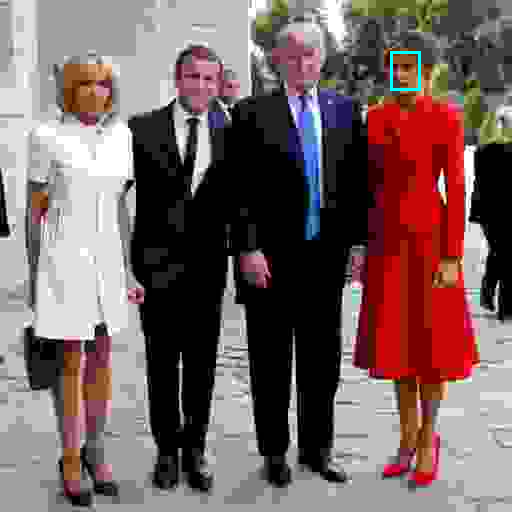}\label{fig:image-comparison-jpeg}}

   \caption{Results of our anonymizing image compression using a region-of-interest loss based on bounding boxes. MTCNN (face) and YOLOv8 (human) detections are annotated in cyan and yellow, respectively. Compression rates are given in bits per pixel. (\subref{fig:image-comparison-original}): original. (\subref{fig:image-comparison-ours}): our method, no face detections, persons in foreground stay detected. (\subref{fig:image-comparison-av1}): AV1 (libaom) at similar bitrate, all four faces detected. (\subref{fig:image-comparison-jpeg}): JPEG at similar bitrate, one face detection, and no human detections.
   }
   \label{fig:image-comparison}
\end{figure}
\par To address this issue, we propose a learned image compression method that anonymizes human faces in the image.
Our learned method employs a convolutional autoencoder that uses a generalized ROI loss function during training.
By using a symmetric autoencoder, the computational complexity of anonymization is effectively split between the edge- and server-side.
Since the analysis transformation into a latent space compresses the input data, we obtain an image codec that depends on both model architecture and the individual weights.
This adds two unknown variables to an attacker without hardware access.
Our approach utilizes the flexibility provided by the parameter sets of the analysis and synthesis transforms to inhibit the reconstruction of human faces while reconstructing other parts.
This way, the extraction of information using other algorithms or neural networks trained on raw image input remains feasible.
Since lossy image compression will inevitably cause a loss of precision of detection methods, we evaluate this deviation by comparing the results of the YOLO \cite{Redmon2016, YOLOv8} object detection approach on the validation dataset of CrowdHuman \cite{shao2018crowdhuman}, by first compressing the image and then feeding it through the detection network.
The bounding boxes of persons returned are then compared to the ground truth using the the average precision (AP) metric of the Pascal VOC Challenge \cite{Everingham2009, Padilla2021}.
Using the same method, we show how our approach leaves faces undetectable, by using bounding-box outputs of the MTCNN face detection network \cite{Zhang2016}.
To set these results into context, we compute the AP of JPEG \cite{Wallace1992} and AV1 \cite{Han2021} at similar compression ratios, while also comparing compression latency and rate, as exemplified in \Cref{fig:image-comparison}.
\par The main contributions of this paper are:
\begin{enumerate}
    \item the combination of a novel ROI-loss function that only operates on bounding boxes and learned compression to achieve a anonymizing compression result,
    \item the evaluation of the change of precision of YOLO and MTCNN caused by anonymization / image compression, and
    \item the comparison of compression latency between our approach and JPEG/AV1, on two different systems.
\end{enumerate}

\section{Related Work}
Classical image compression is a well-studied field of computer science.
Algorithms like JPEG \cite{Wallace1992} and AV1 \cite{Han2021} rely on hand-crafted algorithms to reduce the entropy of the encoded representation.
Recent research shows that learned image compression outperforms non-learned methods in distortion metrics like PSNR and MS-SSIM. \cite{begaint2020compressai}

\subsection{Rate-distortion optimized compression}
One of the first competitive learned approaches was introduced by Ballé \etal. in 2017 \cite{Balle2017}.
It uses layered convolutions for its analysis and synthesis transforms, utilizing the generalized divisive normalization (GDN) nonlinearity \cite{Balle2015} between layers and an uniform quantizer at the bottleneck to achieve coding-efficiency.
In their work, they define a \emph{rate-distortion loss}, which addresses both the reconstruction error of the compression (distortion) and the size of the encoded image (rate), by combining them as a weighted sum.
Since then, improvements to this method have been made partly by improving the entropy model \cite{Balle2018, Cheng2018, Cheng2020}, at the cost of increased computational requirements. 
One recent approach by Liu \etal. \cite{Liu2023} combines layered convolutions with transformers to reach state-of-the-art rate-distortion values.
The idea is that our method is to be used in environments where computational costs for compression cannot rise indefinitely, which renders complex ANN-based compression methods unsuitable.

\subsection{Task-optimized compression}
Classical compression algorithms try to preserve the image quality perceived by humans, not regarding a possible downstream usage of machine vision applications.
Chaiman \etal. \cite{Chamain2021} present end-to-end optimized image compression, where the detection confidence of a downstream algorithm is added to the rate-distortion loss of the learned compression.
Xiao \etal. \cite{Xiao2022} present an identity preserving loss that employs this principle for the task of face recognition, to achieve low compression rates with the model focuses on only reconstructing faces.
Work by Ma \etal. \cite{Ma2021} optimizes perceptual quality by allocating most bits for a given ROI and generating fake texture elsewhere.
Our generalized ROI-loss can also be used to allocate more bits for arbitrary regions in an image.
However, we do not use a ROI network that extracts this foreground for each single compression during preprocessing;
We simply use pre-annotated bounding boxes during training to achieve this behaviour.

\subsection{ANN based anonymization and censorship}
Leaving the field of compression shows that there has already been progress made considering the task of hiding sensitive data in an image. 
One use-case is nudity censorship using adversarial training, as shown by Simões \etal. \cite{Simoes2019} and More \etal. \cite{More2018}.
The task of face anonymization has thus far been achieved by face generation and replacement or inpainting \cite{Maximov2020, Sun2018, Ren2018, Yang2023}.
Most notably however, previous research on face anonymization does not integrate this task into compression.

\par Up to now, literature shows that the main interest lies in optimizing for distortion metrics, as well as some in generative-based anonymization and less in task-optimization. The main standing point of our paper is the combination of both rate-distortion optimized compression and anonymization, as this has not yet been explored by previous research on learned image compression.

\section{Methodology}

\begin{figure*}[t]
    \centering
    \includesvg[width=\linewidth]{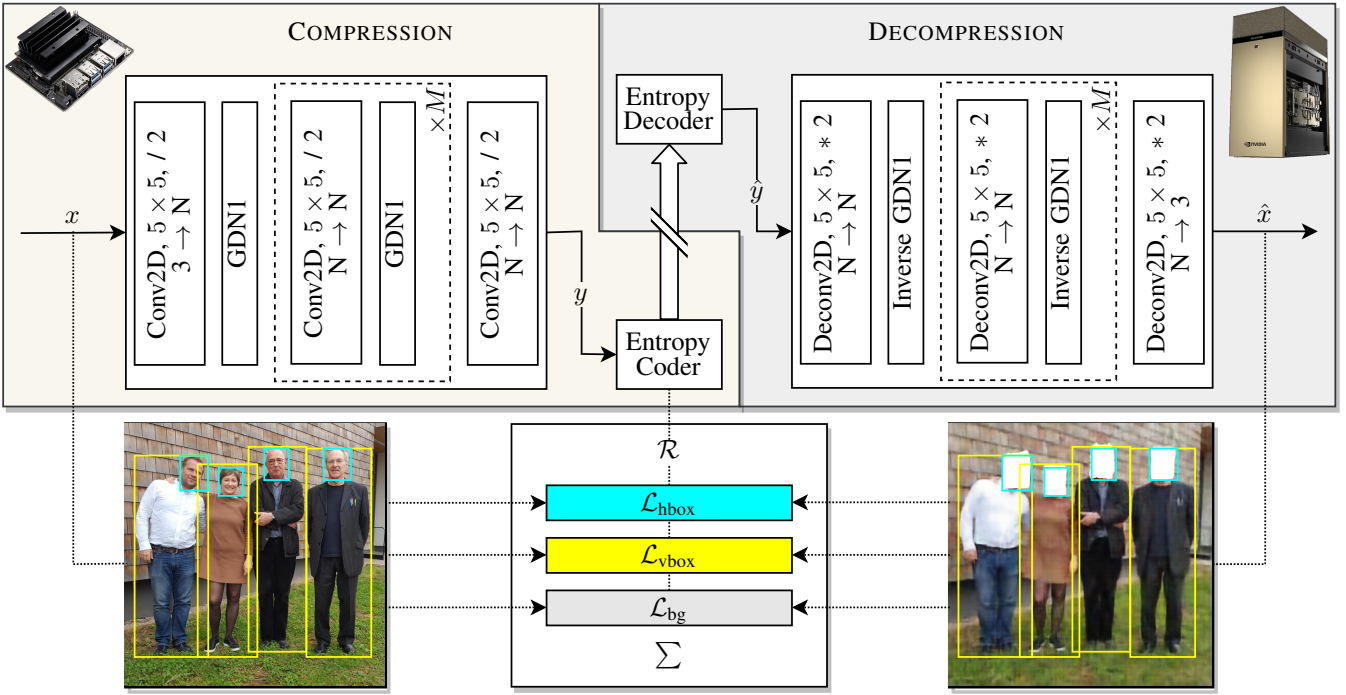}
    \caption{Autoencoder architecture with loss components. Convolutions have kernel size 5 with stride 2, and map the RGB image onto $N$ channels. To assess how AP and bpp vary with different architecture depths and widths, we test different numbers of hidden layers $M\in\left\{1,2\right\}$ and feature maps $N\in\left\{128,256\right\}$. The loss consists of a sum of the compression rate provided by the entropy bottleneck, background-loss and losses for each head-boxes and visual-boxes of persons. (Images of both Jetson Nano and DGX Station A100 were taken from the Nvidia website.)}
    \label{fig:architecture}
\end{figure*}

To achieve compression, one needs to reduce the entropy of a given input.
For this, a nonlinear analysis transform is used first, to reduce the dimensionality of the data by only extracting significant parts. (That is: parts deemed significant by the application.)
An example is the discrete cosine transform of JPEG \cite{Wallace1992}.
Then, the data needs to be quantized and coded into a compact representation, for example, by arithmetic coding.
In order to reconstruct the original image from this compact representation, this process is inverted.
The general approach of learned image compression is replacing the nonlinear transform by a convolutional neural network, and introducing entropy coding at the bottleneck.
This also applies in our case.
Shown in \cref{fig:architecture}, our input image $x$ is analyzed by the encoder to produce a latent representation $y$ that is then quantized and efficiently coded into a representation to be sent over the network channel.
At the receiving end, an entropy decoder reconstructs the latent representation $\hat{y}$ which is then used by the convolutional decoder to synthesize an image $\hat{x}$ that ideally comes close to the original.
\par Now, we introduce the details of our architecture and introduce other parameters present in \cref{fig:architecture}, which are key to the definition of the used loss-function.

\subsection{Architecture}
We use the fully factorized prior architecture introduced by Ballé \etal. in 2018 \cite{Balle2018}. In their work, they use a generalized divisive normalization non-linearity that introduces multiple learnable parameters. 
Johnston \etal. \cite{Johnston2019} argue that by using this non-linearity, unnecessary computational overhead is introduced by computing square roots, while performance-improvements of the model are insignificant.
They continue to simplify this activation by removing parameters that cause this computational overhead, resulting in a simplified GDN.
We use this simplified version to improve latency on the embedded device and denote it by GDN1.
As shown in \cref{fig:architecture}, we vary the depth $M$ and width $N$ of the networks to compare the effects on AP and compression rate.
Each convolutional layer in the analysis (synthesis) transform reduces (increases) both width and height dimensions of the input tensor by two, caused by a step of 2 for each convolution kernel. 
The size of these kernels are fixed to $5\times5$.

\subsection{Training}
During training, we utilize the bounding boxes provided by the CrowdHuman \cite{shao2018crowdhuman} dataset to achieve the anonymization behaviour. This dataset includes pre-annotated bounding boxes for both the head and visible region of persons.
Images are rescaled to $512\times512$ and bounding boxes are adjusted accordingly.
We start defining our loss-function with the rate-distortion components:
\begin{align}
    \mathcal{R}&=\underbrace{\mathbb{E}_{x\sim p_x}\left[-\log_2p_{\hat{y}}(\hat{y})\right]}_{\text{Shannon cross entropy (Rate)}}\\
    \mathcal{L}_{\text{bg}} &= \underbrace{\mathbb{E}_{x\sim p_x}\lVert x - \hat{x} \rVert^2}_{\text{MSE (Distortion)}},
\end{align}
similar to Ballé \etal. \cite{Balle2018}, where $p_x$ is the distribution of images $x$ (during training: a batch),
and $p_{\hat{y}}(\hat{y})$ the entropy-model.
\subsubsection{ROI-Loss: generalized region-of-interest loss}
Now, we introduce a generalized ROI loss that operates on bounding boxes:
\begin{align}
    \mathcal{L}_k(B) = \mathbb{E}_{x\sim p_x, b\in B_x}\left[k - 1^k\cdot\lVert x_b - \hat{x}_b \rVert^2\right],
\end{align}
as the expectancy over the set of bounding boxes $B$ for images of $p_x$. $B_x\subseteq B$ are bounding boxes present in an image $x$, while $x_b$ is an image cropped w.r.t. $b\in B_x$.
For $k=0$, this yields the average loss over given ROI boxes in an image for each image in a batch $p_x$.
Given that inputs and outputs are normalized such that $x,\hat{x}\in\left[0,1\right]$ holds, this loss can be inverted for each box by setting $k=1$, such that close reconstructions compute to a higher loss.
We utilize both variants for boxes provided in the training dataset, non-inverted for persons (vbox; visible bounding box) and inverted for heads (hbox; head bounding box):
\begin{align}
    \mathcal{L}_{\text{vbox}} &= \mathcal{L}_0(\text{vbox}),\\
    \mathcal{L}_{\text{hbox}} &= \mathcal{L}_1(\text{hbox}).
\end{align}
\par The total loss for one batch then computes as a weighted sum over all components:
\begin{align}
    \mathcal{L} = \lambda_r\mathcal{R} + \lambda_{\text{bg}}\mathcal{L}_{\text{bg}} + \lambda_{\text{hbox}}\mathcal{L}_{\text{hbox}} + \lambda_{\text{vbox}}\mathcal{L}_{\text{vbox}}\label{eq:loss-sum}
\end{align}
introducing four new hyperparameters $\lambda_i$ that can be used to prioritize each task.
We use the same hyperparameters for all depth and width configurations, but stop the training of the model with one hidden layer early after 293 epochs.
The two-hidden-layer model was trained up to an epoch of 874, with the narrow variant up to 750.
The hyperparameters are: $\lambda_r=0.04,\:\lambda_{\text{bg}}=1,\:\lambda_{\text{hbox}}=0.6,\:\lambda_{\text{vbox}}=1$.
Although there remains room for optimization on both hyperparameters and depth / width of the architecture, we do not conduct a hyperparameter study in this work.

\section{Evalutaion}

We choose CrowdHuman \cite{shao2018crowdhuman} as the dataset for training and validation.
To evaluate our method, we use both pretrained models YOLOv8 \cite{YOLOv8} and MTCNN \cite{Esler2023} with frozen weights. For YOLOv8, we choose the largest detection model type YOLOv8x for best precision, which is pretrained on the COCO \cite{lin2014microsoft} dataset.
To assess the precision loss by compression, we first apply compression for each image of the CrowdHuman validation dataset (N = 4370), and then run both YOLO and MTCNN detection.
The detected bounding boxes output by YOLO of class person are compared to the visible bounding boxes of CrowdHuman;
In the case of MTCNN, it is the output face bounding boxes against ground truth head bounding boxes.
This will inevitably cause low precision values for MTCNN, since it is not trained on detecting heads;
However since we are only interested in the differences of precision, this is still viable because a certain set of heads in the dataset will show faces.
The same applies for YOLO trained on the COCO dataset, which includes more classes that can be detected.
Here, the drop in precision is lower than for MTCNN.
We believe however that since these foundation models can be considered widespread, it makes sense to evaluate their accuracy instead of the accuracy of some specialized models.
\par We choose the average precision metric of the Pascal VOC challenge \cite{Everingham2009}, the implementation being provided by Padilla \etal{} \cite{Padilla2021}.
This gives us both average precision, as well as the number of true- and false-positives.
In order to compare our results to a baseline, we choose JPEG and AV1 at different quality presets;
Implementations for both codecs are provided by the python imaging library PIL \cite{PIL2023} with AV1 support \cite{Piskun2023}.
\par For a latency assessment, we record the runtime of compression and decompression for every image in the validation dataset, by repeating the process ten times per image and pooling the runtimes over all images, to then compute minimum, arithmetic mean and sample standard deviation.
It is important to note that the compression processes for AV1 run on the CPU, since the support for AV1 encoding is just now starting to be implemented in hardware. 
In our application, the runtime of decoding will happen on a powerful server, while encoding happens on a device with constraints on computing power.
The support for hardware-accelerated AV1 encoding on low-power devices will probably only happen in the far future,
which is why it is reasonable to look at CPU encoding latency for this method.
In contrast, we provide both GPU and CPU measurements for our method where possible. 
Our actual test hardware consists of two devices:
\begin{itemize}
    \item Nvidia DGX Station A100
    \begin{itemize}
        \item 4 x 80GB Nvidia A100 GPU
        \item 64-core AMD EPYC 7742 CPU
    \end{itemize}
    \item Nvidia Jetson Nano
    \begin{itemize}
        \item 4GB Nvidia Maxwell GPU
        \item Quad-core ARM Cortex-A57. (We skip benchmarking our method on this CPU.)
    \end{itemize}
\end{itemize}
When using GPU computations, we do not rescale to half-precision floating point, nor decrease the accuracy of matrix calculations. 
In case of the DGX station, we use only one GPU for latency measurements.

\section{Results}
\subsection{Precision}
\begin{figure*}[t]
    \centering%
    \subfloat[YOLOv8 person detection, linear AP scale.]{\includesvg[width=.5\linewidth]{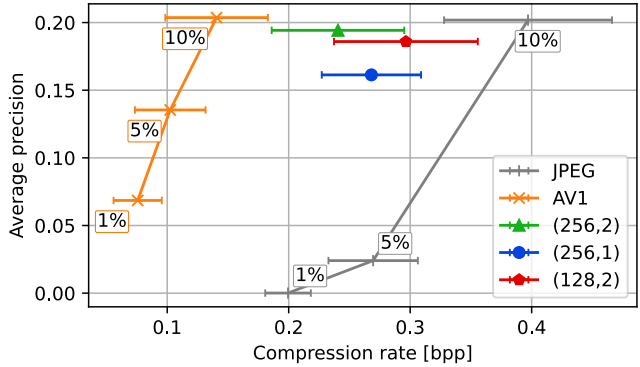}\label{fig:precision-person}}%
    \subfloat[MTCNN face detection, logarithmic AP scale.]{\includesvg[width=.5\linewidth]{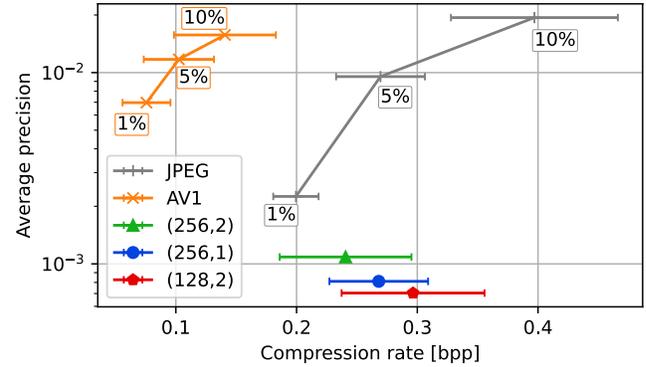}\label{fig:precision-face}}%
    \caption{Average precision over compression rate for JPEG, AV1 and our method with three configurations, at an IOU-threshold of 0.5. For JPEG and AV1, we choose three quality presets: 1, 5 and 10 percent. Error-bars indicate one sample standard deviation of compression rate over the validation dataset. Subfigure \subref{fig:precision-person} shows the average precision of YOLOv8 on persons, \subref{fig:precision-face} the average precision of MTCNN on faces; Both after compression.}
    \label{fig:precision-classes}
\end{figure*}

\cref{fig:precision-classes} shows visualized results of average precision over compression rate, where \cref{fig:precision-person} are the results for YOLOv8x and \cref{fig:precision-face} for MTCNN.
We include JPEG and AV1 at three quality presets, and our method at three configurations: $(N,M)\in\left\{(128,2),(256,1),(256,2)\right\}$.
Note that there is only one value for each configuration, since our method does not facilitate setting a quality preset at runtime;
This precision-distortion trade-off is to be adjusted during training with the hyperparameters $\lambda_i$ shown in \cref{eq:loss-sum}.
Our method with the configuration $(256,2)$ achieves similar precision of 19\% AP on persons to AV1 at 10\% quality (20\% AP, 0.14 bpp) and JPEG at 10\% quality (20\% AP, 0.39 bpp), at a bitrate of 0.24 bits per pixel.
Removing one hidden layer $(256,1)$ decreases the precision to 16\% AP and increases the bitrate to 0.26 bits per pixel.
Reducing the width $(128,2)$ requires a higher bitrate to achieve similar precision values to $(256,2)$.
The precision on heads / faces of two hidden layers (0.1\% AP) is seven times lower in comparison to 0.7\% AP of AV1 at 1\% quality, while one hidden layer reduces this precision to 0.08\% AP at a similar bitrate to JPEG at 5\% quality with a precision of 0.9\% AP.
Reducing the width of the network to 128 and training up to 750 epochs yields 0.07\% AP on faces.
\par Considering true and false positive detections, \cref{tab:detection-positives} shows that using $(128,2)$ as a model configuration decreases the number of true positives to 150 in comparison to the lowest achieved rate of JPEG at 1\% quality of 424 true positives.
\begin{table}[t]
    \centering%
    \caption{\textsc{Number of true and false positive face detections.} Ours (best in total) in bold.}\label{tab:detection-positives}%
    \begin{tabularx}{\linewidth}{
  |>{\centering\arraybackslash}X
  |>{\centering\arraybackslash}X
  |>{\centering\arraybackslash}X
  |>{\centering\arraybackslash}X|}
         \multicolumn{1}{c}{Method} & \multicolumn{1}{c}{Preset} & \multicolumn{1}{c}{TP} & \multicolumn{1}{c}{FP} \\\hline\hline%
         $(256,2)$ & \multirow{3}{*}{\diagbox[innerwidth=6em, height=3\line]{}{}} & 234 & 1159 \\\cline{1-1}\cline{3-4}%
         $(256,1)$ & & 191 & 1488 \\\cline{1-1}\cline{3-4}%
         $(128,2)$ & & \textbf{150} & 1105 \\\hline%
         \multirow{3}{*}{AV1} & 1\% & 1408 & 1949 \\\cline{2-4}%
          & 5\% & 2253 & 2966 \\\cline{2-4}%
          & 10\% & 3138 & 4345 \\\hline%
         \multirow{3}{*}{JPEG} & 1\% & 424 & 613 \\\cline{2-4}%
          & 5\% & 1995 & 3180 \\\cline{2-4}%
          & 10\% & 4031 & 6399 \\\hline%
    \end{tabularx}
\end{table}

\subsection{Latency} 
\begin{figure*}[t]
    \centering%
    \subfloat[Jetson Nano]{\includesvg[width=.5\linewidth]{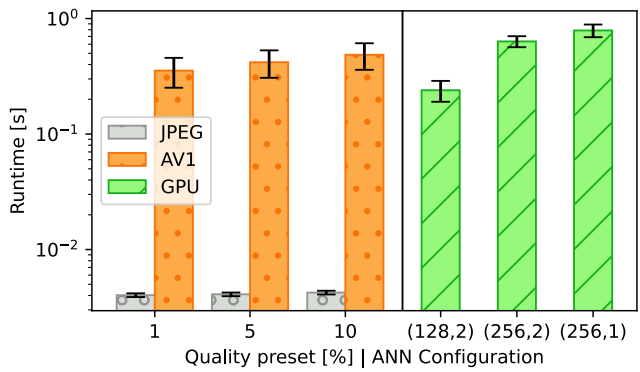}\label{fig:latency-jetson}}%
    \subfloat[DGX station]{\includesvg[width=.5\linewidth]{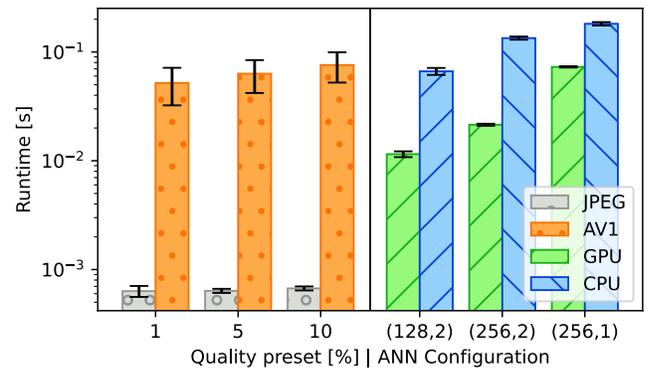}\label{fig:latency-a100}}%
    \caption{Encoding latency comparison of JPEG and AV1 at 1, 5 and 10 to the left, against our method with three configurations to the right. Error-bars indicate one sample standard deviation in latency over the validation dataset. Logarithmic runtime scale.}
    \label{fig:latency}
\end{figure*}

\cref{fig:latency} compares each method on each of our devices in encoding latency. For further reference, numeric values are provided in \cref{tab:latency-jetson} and \cref{tab:latency-a100}.
On the Nvidia DGX station, our encoding method achieves lower latency than AV1 at 1\% quality at 52 ms, with the configuration $(256,2)$ while running on the GPU at 21 ms.
Decreasing the depth of the model does not reduce the runtime of the model, while reducing width does significantly;
Increasing the runtime three times and reducing it by a factor of two, respectively.
On the Jetson Nano, runtimes of both $(256,2),(256,1)$ configurations do not outperform AV1.
The reduced-width of $(128,2)$ decreases the latency to 239 ms, resulting in 110 ms less than minimum AV1.

\begin{table}[t]
\centering
\caption{\textsc{Jetson Nano: encoding latency.} Values of \Cref{fig:latency-jetson}. Lowest latency in italics. Best of ours in bold.}\label{tab:latency-jetson}
\begin{tabularx}{\linewidth} { 
  |>{\hsize=.75\hsize\centering\arraybackslash}X 
  |>{\hsize=.75\hsize\centering\arraybackslash}X
  |>{\hsize=1.5\hsize\centering\arraybackslash}X
  |>{\centering\arraybackslash}X
  |>{\centering\arraybackslash}X|}
 \multicolumn{1}{c}{Method} & \multicolumn{1}{c}{Preset} & \multicolumn{1}{c}{$\overline{t} \pm s_t$} & \multicolumn{1}{c}{$\min t$} & \multicolumn{1}{c}{Unit} \\\hline\hline
 $(256,2)$ & \multirow{3}{*}{GPU} & $0.634\pm0.069$ & $0.610$ & \multirow{6}{*}{[s]} \\\cline{1-1}\cline{3-4}
 $(256,1)$ & & $0.789\pm0.010$ & $0.755$ & \\\cline{1-1}\cline{3-4}
 $(128,2)$ & & $\mathbf{0.239\pm0.049}$ & $0.221$ & \\\cline{1-4}
 \multirow{3}{*}{AV1} & 1\% & $0.35\pm0.10$ & $0.18$ & \\\cline{2-4}
  & 5\% & $0.42\pm0.11$ & $0.20$ & \\\cline{2-4}
  & 10\% & $0.49\pm0.13$ & $0.22$ & \\\hline
 \multirow{3}{*}{JPEG} & 1\% & $\mathit{4.03\pm0.15}$ & $3.72$ & \multirow{3}{*}{[ms]} \\\cline{2-4}
  & 5\% & $4.10\pm0.15$ & $3.76$ & \\\cline{2-4}
  & 10\% & $4.24\pm0.16$ & $3.79$ & \\\hline
\end{tabularx}
\end{table}
\begin{table}[t]
\centering
\caption{\textsc{DGX Station: encoding latency.} Values of \Cref{fig:latency-a100}. Lowest latency in italics. Best of ours in bold.}\label{tab:latency-a100}
\begin{tabularx}{\linewidth} {
  |>{\hsize=.75\hsize\centering\arraybackslash}X 
  |>{\hsize=.75\hsize\centering\arraybackslash}X
  |>{\hsize=1.5\hsize\centering\arraybackslash}X
  |>{\centering\arraybackslash}X
  |>{\centering\arraybackslash}X|}
 \multicolumn{1}{c}{Method} & \multicolumn{1}{c}{Preset} & \multicolumn{1}{c}{$\overline{t} \pm s_t$} & \multicolumn{1}{c}{$\min t$} & \multicolumn{1}{c}{Unit} \\\hline\hline
 \multirow{2}{*}{$(256,2)$} & GPU & $21.43\pm0.46$ & $20.81$ & \multirow{9}{*}{[ms]}\\\cline{2-4}
 & CPU & $134.1\pm4.5$ & $111.3$ & \\\cline{1-4}
 \multirow{2}{*}{$(256,1)$} & GPU & $72.9\pm1.0$ & $70.7$ & \\\cline{2-4}
 & CPU & $181.6\pm6.6$ & $147.5$ & \\\cline{1-4}
 \multirow{2}{*}{$(128,2)$} & GPU & $\mathbf{11.50\pm0.71}$ & $10.41$ & \\\cline{2-4}
 & CPU & $66.4\pm4.9$ & $40.9$ & \\\cline{1-4}
 \multirow{3}{*}{AV1} & 1\% & $52\pm20$ & $20$ & \\\cline{2-4}
  & 5\% & $64\pm21$ & $25$ & \\\cline{2-4}
  & 10\% & $76\pm24$ & $29$ & \\\hline
 \multirow{3}{*}{JPEG} & 1\% & $\mathit{633\pm73}$ & $564$ & \multirow{3}{*}{[µs]}\\\cline{2-4}
  & 5\% & $637\pm28$ & $570$ & \\\cline{2-4}
  & 10\% & $671\pm29$ & $577$ & \\\hline
\end{tabularx}
\end{table}

\section{Discussion}

The results show that anonymization behavior can be learned using less-complex architectures.
Reducing the number of available parameters decreases downstream people detection precision at similar bit-rates, while providing latency benefits in case of networks with reduced width but unchanged depth.
\Cref{tab:detection-positives} shows that while our method successfully reduces the number of true positive face detections down to 150 over the whole test dataset, the number of true positive face detections is not zero.
The reason for this are large faces, of which the training dataset includes only few.
In comparison with JPEG, we achieve a better precision-rate trade-off across all of our configurations.
Comparing to AV1 shows that ANN-based compression is able to outperform it on latency.
\par While standardized codecs excel in many areas, they fall short in providing application flexibility.
One notable example of this limitation is the inability to anonymize faces during the encoding and decoding process.
We especially argue that the combination of both compression and anonymization comes with benefits that other approaches cannot provide;
Considering the handling of sensitive data, using our method alleviates any distrust present in the party tasked with decompression, since the original image data is already altered during encoding.
Our method also renders man-in-the-middle attacks meaningless, since access to the encoded data requires knowledge of both decoder architecture and weights to reconstruct the image data.
Lastly, by combining anonymization and compression, we show that it is unnecessary to chain standardized codecs and high-complexity ANN-based anonymization techniques introduced by previous research.
This especially benefits the use of our method in edge-devices, since a part of the process happens on another device tasked with decoding.
\par Low-bitrate compression on lower-power devices remains a rarely approached topic.
In this field, learned image compression is a competitive alternative, especially when dedicated hardware encoding implementations are unavailable.
On lower-power devices that have access to a dedicated GPU or TPU, the advantage of using ANN-based compression is even greater, as these accelerators further decrease latency.
Furthermore, the implementation of more generalized matrix extensions like those implemented by Intel \cite{Intel2023} and presented by ARM \cite{ARMCommunity2023} into lower-power CPUs will make the usage of methods like ours more feasible, even when a dedicated accelerator is unavailable.
\par Additionally, the application of our loss-function operating on bounding boxes is not limited to face anonymization;
It can easily be generalized by using it with any dataset that comes with bounding boxes.
This opens up the possibility to hide or accentuate any kind of region of interest in images, like censoring nudity or highlighting weapons.

\section{Conclusion}

In this paper, we show how using a region-of-interest loss operating on pre-annotated bounding boxes can be used to utilize the flexibility of ANN based compression methods to highlight or hide specific parts in an image.
As an application, we introduce the task of face anonymization.
Our network learns to in-paint faces in a given image, making them undetectable in the decoded result, while keeping persons detectable.
We show that this detectability of humans is kept possible while faces become unrecognizable by comparing the detection results of the foundation models YOLOv8 (people) and MTCNN (faces) against the pre-annotated ground truth of the CrowdHuman dataset.
Comparing these results to widespread JPEG and state-of-the-art AV1-based image compression as a baseline, we show that our method achieves much higher precision on persons than JPEG at similar bitrates, while keeping face detections at an absolute minimum.
From a latency perspective, our method outperforms AV1 on encoding while utilizing the flexibility of the ANN network for the additional step of anonymization.
Additionally, we discuss the implications of the usage of our solution in edge computing environments.


\bibliographystyle{IEEEtran.bst}
\bibliography{egbib}

\begin{thebibliography}{10}
\providecommand{\url}[1]{#1}
\csname url@rmstyle\endcsname
\providecommand{\newblock}{\relax}
\providecommand{\bibinfo}[2]{#2}
\providecommand\BIBentrySTDinterwordspacing{\spaceskip=0pt\relax}
\providecommand\BIBentryALTinterwordstretchfactor{4}
\providecommand\BIBentryALTinterwordspacing{\spaceskip=\fontdimen2\font plus
\BIBentryALTinterwordstretchfactor\fontdimen3\font minus \fontdimen4\font\relax}
\providecommand\BIBforeignlanguage[2]{{%
\expandafter\ifx\csname l@#1\endcsname\relax
\typeout{** WARNING: IEEEtran.bst: No hyphenation pattern has been}%
\typeout{** loaded for the language `#1'. Using the pattern for}%
\typeout{** the default language instead.}%
\else
\language=\csname l@#1\endcsname
\fi
#2}}

\bibitem{EU2023}
\BIBentryALTinterwordspacing
{European Parliamentary Research Service}, ``{EU Artificial intelligence act},'' June 2023. [Online]. Available: \url{https://www.europarl.europa.eu/RegData/etudes/BRIE/2021/698792/EPRS_BRI(2021)698792_EN.pdf}
\BIBentrySTDinterwordspacing

\bibitem{Yang2023}
J.~Yang, S.~Qiao, Z.~Wang, and Z.~Zuo, ``{Adversarial Secret-Identity Generation Model for Face Anonymization in the Internet of Vehicles},'' \emph{IEEE Systems Journal}, pp. 1--10, 2023.

\bibitem{More2018}
M.~D. More, D.~M. Souza, J.~Wehrmann, and R.~C. Barros, ``{Seamless Nudity Censorship: an Image-to-Image Translation Approach based on Adversarial Training},'' in \emph{{2018 International Joint Conference on Neural Networks (IJCNN)}}, 2018, pp. 1--8.

\bibitem{Redmon2016}
J.~Redmon, S.~Divvala, R.~Girshick, and A.~Farhadi, ``{You Only Look Once: Unified, Real-Time Object Detection},'' in \emph{{Proceedings of the IEEE Conference on Computer Vision and Pattern Recognition (CVPR)}}, June 2016.

\bibitem{YOLOv8}
\BIBentryALTinterwordspacing
G.~Jocher, A.~Chaurasia, and J.~Qiu, ``{YOLO by Ultralytics},'' Jan. 2023. [Online]. Available: \url{https://github.com/ultralytics/ultralytics}
\BIBentrySTDinterwordspacing

\bibitem{shao2018crowdhuman}
S.~Shao, Z.~Zhao, B.~Li, T.~Xiao, G.~Yu, X.~Zhang, and J.~Sun, ``{CrowdHuman: A Benchmark for Detecting Human in a Crowd},'' \emph{arXiv preprint arXiv:1805.00123}, 2018.

\bibitem{Everingham2009}
M.~Everingham, L.~V. Gool, C.~K.~I. Williams, J.~Winn, and A.~Zisserman, ``The pascal visual object classes ({VOC}) challenge,'' \emph{International Journal of Computer Vision}, vol.~88, no.~2, pp. 303--338, sep 2009.

\bibitem{Padilla2021}
R.~Padilla, W.~L. Passos, T.~L.~B. Dias, S.~L. Netto, and E.~A.~B. da~Silva, ``A comparative analysis of object detection metrics with a companion open-source toolkit,'' \emph{Electronics}, vol.~10, no.~3, p. 279, jan 2021.

\bibitem{Zhang2016}
\BIBentryALTinterwordspacing
K.~Zhang, Z.~Zhang, Z.~Li, and Y.~Qiao, ``Joint {Face} {Detection} and {Alignment} using {Multi}-task {Cascaded} {Convolutional} {Networks},'' \emph{IEEE Signal Processing Letters}, vol.~23, no.~10, pp. 1499--1503, Oct. 2016, arXiv:1604.02878 [cs]. [Online]. Available: \url{http://arxiv.org/abs/1604.02878}
\BIBentrySTDinterwordspacing

\bibitem{Wallace1992}
G.~Wallace, ``The {JPEG} still picture compression standard,'' \emph{IEEE Transactions on Consumer Electronics}, vol.~38, no.~1, pp. xviii--xxxiv, Feb. 1992.

\bibitem{Han2021}
J.~Han, B.~Li, D.~Mukherjee, C.-H. Chiang, A.~Grange, C.~Chen, H.~Su, S.~Parker, S.~Deng, U.~Joshi, Y.~Chen, Y.~Wang, P.~Wilkins, Y.~Xu, and J.~Bankoski, ``A technical overview of av1,'' \emph{Proceedings of the IEEE}, vol. 109, no.~9, pp. 1435--1462, 2021.

\bibitem{begaint2020compressai}
J.~Bégaint, F.~Racapé, S.~Feltman, and A.~Pushparaja, ``{CompressAI: a PyTorch library and evaluation platform for end-to-end compression research},'' 2020.

\bibitem{Balle2017}
\BIBentryALTinterwordspacing
J.~Ballé, V.~Laparra, and E.~P. Simoncelli, ``End-to-end optimized image compression,'' in \emph{International Conference on Learning Representations}, 2017. [Online]. Available: \url{https://openreview.net/forum?id=rJxdQ3jeg}
\BIBentrySTDinterwordspacing

\bibitem{Balle2015}
------, ``Density modeling of images using a generalized normalization transformation,'' \emph{Int'l Conf on Learning Representations (ICLR), San Juan, Puerto Rico, May 2016}, Nov. 2015.

\bibitem{Balle2018}
\BIBentryALTinterwordspacing
J.~Ballé, D.~Minnen, S.~Singh, S.~J. Hwang, and N.~Johnston, ``Variational image compression with a scale hyperprior,'' in \emph{{International Conference on Learning Representations}}, 2018. [Online]. Available: \url{https://openreview.net/forum?id=rkcQFMZRb}
\BIBentrySTDinterwordspacing

\bibitem{Cheng2018}
Z.~Cheng, H.~Sun, M.~Takeuchi, and J.~Katto, ``Deep convolutional autoencoder-based lossy image compression,'' in \emph{2018 Picture Coding Symposium (PCS)}, 2018, pp. 253--257.

\bibitem{Cheng2020}
------, ``{Learned Image Compression With Discretized Gaussian Mixture Likelihoods and Attention Modules},'' in \emph{{Proceedings of the IEEE/CVF Conference on Computer Vision and Pattern Recognition (CVPR)}}, June 2020.

\bibitem{Liu2023}
J.~Liu, H.~Sun, and J.~Katto, ``{Learned Image Compression With Mixed Transformer-CNN Architectures},'' in \emph{{Proceedings of the IEEE/CVF Conference on Computer Vision and Pattern Recognition (CVPR)}}, June 2023, pp. 14\,388--14\,397.

\bibitem{Chamain2021}
L.~D. Chamain, F.~Racapé, J.~Bégaint, A.~Pushparaja, and S.~Feltman, ``End-to-{End} optimized image compression for machines, a study,'' in \emph{2021 {Data} {Compression} {Conference} ({DCC})}, Mar. 2021, pp. 163--172, iSSN: 2375-0359.

\bibitem{Xiao2022}
J.~Xiao, L.~Aggarwal, P.~Banerjee, M.~Aggarwal, and G.~Medioni, ``{Identity Preserving Loss for Learned Image Compression},'' in \emph{{Proceedings of the IEEE/CVF Conference on Computer Vision and Pattern Recognition (CVPR) Workshops}}, June 2022, pp. 517--526.

\bibitem{Ma2021}
Y.~Ma, Y.~Zhai, C.~Yang, J.~Yang, R.~Wang, J.~Zhou, K.~Li, Y.~Chen, and R.~Wang, ``{Variable Rate ROI Image Compression Optimized for Visual Quality},'' in \emph{{Proceedings of the IEEE/CVF Conference on Computer Vision and Pattern Recognition (CVPR) Workshops}}, June 2021, pp. 1936--1940.

\bibitem{Simoes2019}
G.~S. Simões, J.~Wehrmann, and R.~C. Barros, ``{Attention-based Adversarial Training for Seamless Nudity Censorship},'' in \emph{{2019 International Joint Conference on Neural Networks (IJCNN)}}, 2019, pp. 1--8.

\bibitem{Maximov2020}
M.~Maximov, I.~Elezi, and L.~Leal-Taixe, ``{CIAGAN: Conditional Identity Anonymization Generative Adversarial Networks},'' in \emph{{Proceedings of the IEEE/CVF Conference on Computer Vision and Pattern Recognition (CVPR)}}, June 2020.

\bibitem{Sun2018}
Q.~Sun, L.~Ma, S.~J. Oh, L.~Van~Gool, B.~Schiele, and M.~Fritz, ``{Natural and Effective Obfuscation by Head Inpainting},'' in \emph{{Proceedings of the IEEE Conference on Computer Vision and Pattern Recognition (CVPR)}}, June 2018.

\bibitem{Ren2018}
Z.~Ren, Y.~J. Lee, and M.~S. Ryoo, ``{Learning to Anonymize Faces for Privacy Preserving Action Detection},'' in \emph{{Proceedings of the European Conference on Computer Vision (ECCV)}}, September 2018.

\bibitem{Johnston2019}
N.~Johnston, E.~Eban, A.~Gordon, and J.~Ballé, ``{Computationally Efficient Neural Image Compression},'' 2019.

\bibitem{Esler2023}
\BIBentryALTinterwordspacing
T.~Esler, ``Face {Recognition} {Using} {Pytorch},'' July 2023, original-date: 2019-05-25T01:29:24Z. [Online]. Available: \url{https://github.com/timesler/facenet-pytorch}
\BIBentrySTDinterwordspacing

\bibitem{lin2014microsoft}
T.-Y. Lin, M.~Maire, S.~Belongie, J.~Hays, P.~Perona, D.~Ramanan, P.~Doll{\'a}r, and C.~L. Zitnick, ``{Microsoft coco: Common objects in context},'' in \emph{{Computer Vision--ECCV 2014: 13th European Conference, Zurich, Switzerland, September 6-12, 2014, Proceedings, Part V 13}}.\hskip 1em plus 0.5em minus 0.4em\relax Springer, 2014, pp. 740--755.

\bibitem{PIL2023}
\BIBentryALTinterwordspacing
{PIL Contributors}, ``{Pillow Imaging Library (Fork)},'' July 2023, original-date: 2012-07-24T21:38:39Z. [Online]. Available: \url{https://github.com/python-pillow/Pillow}
\BIBentrySTDinterwordspacing

\bibitem{Piskun2023}
\BIBentryALTinterwordspacing
A.~Piskun, ``pillow-heif,'' July 2023, original-date: 2021-09-10T16:24:22Z. [Online]. Available: \url{https://github.com/bigcat88/pillow_heif}
\BIBentrySTDinterwordspacing

\bibitem{Intel2023}
\BIBentryALTinterwordspacing
Intel, ``\BIBforeignlanguage{en}{Intel® {Advanced} {Matrix} {Extensions} {Overview}}.'' [Online]. Available: \url{https://www.intel.com/content/www/us/en/products/docs/accelerator-engines/advanced-matrix-extensions/overview.html}
\BIBentrySTDinterwordspacing

\bibitem{ARMCommunity2023}
\BIBentryALTinterwordspacing
M.~Weidmann, ``\BIBforeignlanguage{en}{{Introducing the Scalable Matrix Extension for the Armv9-A Architecture}}.'' [Online]. Available: \url{https://community.arm.com/arm-community-blogs/b/architectures-and-processors-blog/posts/scalable-matrix-extension-armv9-a-architecture}
\BIBentrySTDinterwordspacing

\end{thebibliography}

\end{document}